\documentclass{article} 
\usepackage{nips15submit_e}

\usepackage[T1]{fontenc}

\usepackage{mathptmx}
\usepackage{amsmath}
\usepackage{amsthm}
\usepackage{amssymb}

\usepackage[final,expansion=alltext]{microtype}     
\usepackage{graphicx}                               
\usepackage{amsmath}                                
\usepackage[labelfont=bf]{caption}                  
\usepackage[format=hang]{subcaption}                
\usepackage{booktabs}                               
\usepackage[numbers,sort]{natbib}                   
\usepackage[acronym,smallcaps,nowarn]{glossaries}   

\DeclareRobustCommand{\parhead}[1]{\textbf{#1} }

\newcommand{\E}{\mathbb{E}}

\newcommand{\cL}{\mathcal{L}}

\newacronym{KL}{kl}{Kullback-Leibler}
\newacronym{ELBO}{elbo}{evidence lower bound}

\newacronym{SVI}{svi}{stochastic variational inference}

\newacronym{AD}{ad}{automatic differentiation}

\newacronym{VBSTAN}{vbstan}{variational Bayes in Stan}

\newacronym{ARD}{ard}{automatic relevance determination}

\newacronym{EP}{ep}{expectation propagation}
\newacronym{EPA}{epa}{expectation propagation algorithm}

\usepackage{color,soul}
\usepackage{algorithm,algorithmic}
\usepackage{bibunits}

\newcommand\g[1][]{\:#1\vert\:}

\newcommand{\bx}{\mathbf{x}}
\newcommand{\bX}{\mathbf{X}}
\newcommand{\bz}{\mathbf{z}}
\newcommand{\bZ}{\mathbf{Z}}

\title{The Population Posterior \\ and Bayesian Inference on Streams}

\author{
James McInerney \\
Data Science Institute\\
Columbia University\\
New York, NY 10027 \\
\texttt{jm4181@columbia.edu} \\
\And
Rajesh Ranganath\\
Computer Science Department\\
Princeton University\\
Princeton, NJ 08540 \\
\texttt{rajeshr@cs.princeton.edu} \\
\And
David M. Blei \\
Data Science Institute, Department of Statistics \&\\ Department of Computer Science\\
Columbia University\\
New York, NY 10027 \\
\texttt{david.blei@columbia.edu}
}

\nipsfinalcopy 

\begin{document}

\maketitle

\begin{abstract}
Many modern data analysis problems involve inferences from streaming data. However, streaming data is not easily amenable to the standard probabilistic modeling approaches, which assume that we condition on finite data. We develop population variational Bayes, a new approach for using Bayesian modeling to analyze streams of data. It approximates a new type of distribution, the population posterior, which combines the notion of a population distribution of the data with Bayesian inference in a probabilistic model.  We study our method with latent Dirichlet allocation and Dirichlet process mixtures on several large-scale data sets.

\end{abstract}

\section{Introduction}

Probabilistic modeling has emerged as a powerful tool for data
analysis. It is an intuitive language for describing assumptions about
data and provides efficient algorithms for analyzing real data under
those assumptions. The main idea comes from Bayesian statistics. We
encode our assumptions about the data in a structured probability
model of hidden and observed variables; we condition on a data set to
reveal the posterior distribution of the hidden variables; and we use
the resulting posterior as needed, for example to form predictions
through the posterior predictive distribution or to explore the data
through the posterior expectations of the hidden variables.

Many modern data analysis problems involve inferences from streaming
data. Examples include exploring the content of massive social media streams (e.g., Twitter, Facebook), 
analyzing live video streams, 
modeling the preferences of users on an online platform for recommending new items, 
and predicting human mobility patterns for anticipatory computing. 
Such problems, however, cannot easily take
advantage of the standard approach to probabilistic modeling, which
always assumes that we condition on a finite data set. This might be
surprising to some readers; after all, one of the tenets of the
Bayesian paradigm is that we can update our posterior when given new
information. (``Yesterday's posterior is today's prior.''). But there
are two problems with using Bayesian updating on data streams.

The first problem is that Bayesian posteriors will become
overconfident. Conditional on never-ending data, most posterior
distributions (with a few exceptions) result in a point mass at a
single configuration of the latent variables. That is, the posterior
contains no variance around its idea of the hidden variables that
generated the observed data. In theory this is sensible, but only in
the impossible scenario where the data truly came from the proposed
model. In practice, all models provide approximations to the
data-generating distribution, and we always harbor uncertainty about
the model (and thus the hidden variables) even in the face of an
infinite data stream.

The second problem is that the data stream might change over time.
This is an issue because, frequently, our goal in applying
probabilistic models to streams is not to characterize how they
change, but rather to accommodate it. That is, we would like for our
current estimate of the latent variables to be accurate to the current
state of the stream and to adapt to how the stream might slowly
change. (This is in contrast, for example, to time series modeling.)
Traditional Bayesian updating cannot handle this. Either we explicitly
model the time series, and pay a heavy inferential cost, or we tacitly
assume that the data are independent and exchangeable.

In this paper we develop new ideas for analyzing data streams with
probabilistic models. Our approach combines the frequentist notion of
the population distribution with probabilistic models and Bayesian
inference.

\parhead{Main idea: The population posterior.} Consider a latent
variable model of $\alpha$ data points. (This is unconventional
notation; we will describe why we use it below.)
Following~\cite{hoffman_stochastic_2013}, we define the model to have
two kinds of hidden variables: global hidden variables $\beta$ contain
latent structure that potentially governs any data point; local hidden
variables $z_i$ contain latent structure that only governs the $i$th
data point. Such models are defined by the joint, \begin{align}
  \label{eq:joint}
  p(\beta, \mathbf{z}, \mathbf{x}) = p(\beta) \prod_{i=1}^{\alpha}
  p(x_i, z_i \g \beta),
\end{align}
where $\bx = x_{1:\alpha}$ and $\bz = z_{1:\alpha}$. Traditional
Bayesian statistics conditions on a fixed data set $\bx$ to obtain the
posterior distribution of the hidden variables $p(\beta, \bz \g \bx)$.
As we discussed, this framework cannot accommodate data streams. We
need a different way to use the model.

We define a new distribution, the \textit{population posterior}, which
enables us to consider Bayesian modeling of streams. Suppose we
observe $\alpha$ data points independently from the underlying
population distribution, $\bX \sim F_{\alpha}$. This induces a
posterior $p(\beta, \bz \g \bX)$, which is a function of the random
data. The expected value of this posterior distribution is the
population posterior, \begin{align}
  \E_{F_\alpha}\left[p(\bz, \beta | \bX) \right] =
  \E_{F_\alpha} \left[ \frac{p(\beta, \bz, \bX)}{p(\bX)} \right].
  \label{eq:pop_post}
\end{align}
Notice that this distribution is not a function of observed data; it
is a function of the population distribution $F$ and the data set size
$\alpha$. The data set size is a parameter that can be set. 
This parameter controls the variance of the population posterior, and so depends on how close the model is to the true data distribution.

We have defined a new problem. Given an endless stream of data points
coming from $F$ and a value for $\alpha$, our goal is to approximate
the corresponding population posterior. We will use variational
inference and stochastic optimization to approximate the population
posterior. As we will show, our algorithm justifies applying a variant
of stochastic variational inference~\citep{hoffman_stochastic_2013} to
a data stream. We used this technique to analyze several data streams
with modern probabilistic models, such as latent Dirichlet allocation
and Dirichlet process mixtures. With held out likelihood as a measure
of model fitness, we found our method to give better models of the
data than approaches based on full Bayesian
inference~\citep{hoffman_stochastic_2013} or Bayesian
updating~\citep{broderick_streaming_2013}.

\parhead{Related work.} Several methods exist for performing inference
on streams of data. Refs.~\citep{yao_efficient_2009,
  ahmed_online_2011, doucet_sequential_2000}~propose extending Markov
chain Monte Carlo methods for streaming data. However, sampling-based
approaches do not scale to massive datasets. The variational
approximation enables more scalable
inference. Ref.~\cite{honkela_online_2003}~propose online variational
inference by exponentially forgetting the variational parameters
associated with old data. Ref.~\cite{hoffman_stochastic_2013} also
decay parameters derived from old data, but interpret this action in
the context of stochastic optimization, bringing guarantees of
convergence to a local optimum.  This gradient-based approach has
enabled the application of more advanced probabilistic models to
large-scale data sets.  However, none of these methods are applicable
to streaming data, because they implicitly rely on the data being of
known size (even when based on subsampling data points to obtain noisy
gradients).

To apply the variational approximation to streaming data,
Ref.~\cite{broderick_streaming_2013} and
Ref.~\cite{ghahramani_online_2000} both propose performing Bayesian
updating to the approximating family.  Their method uses the latest
approximation having seen $n$ data points as a prior to be updated
using the approximation of the next data point (or mini-batch) using
Bayes' rule. Ref.~\citep{tank_streaming_2015} adapt this framework to
nonparametric mixture modelling.  Here we take a different approach,
by changing the overall variational objective to incorporate a
population distribution $F$ and following stochastic gradients of this
new objective.

Independently, Ref.~\cite{theis_trust_2015} apply SVI to streaming
settings by accumulating new data points into a growing window, then
uniformly sampling from this window to update the variational
parameters.  Our method justifies that approach. Further, they propose
updating parameters along a trust region, instead of following
(natural) gradients, as a way of mitigating local optima.  This
innovation can be incorporated into our method.

\section{Variational Inference for the Population Posterior}

We develop \textit{population variational Bayes}, a method for
approximating the population posterior in Eq.~\ref{eq:pop_post}.  Our
method is based on variational inference and stochastic optimization.

\parhead{The F-ELBO.}  The idea behind variational inference is to
approximate difficult-to-compute distributions through
optimization~\cite{jordan_introduction_1999,wainwright_graphical_2008}.
We introduce an approximating family of distributions over the latent
variables $q(\beta, \bz)$ and try to find the member of $q(\cdot)$
that minimizes the Kullback-Leibler (KL) divergence to the target
distribution.

Population variational Bayes (VB) uses variational inference to
approximate the population posterior in Eq.~\ref{eq:pop_post}.
It aims to solve the following problem,
\begin{align}
  q^*(\beta, \bz) = \min_q \mathrm{KL}(q(\beta, \bz) ||
  \E_{F_\alpha}[p(\beta, \bz \g \bX)]).
  \label{eq:kl}
\end{align}
As for the population posterior, this objective is a function of the
population distribution of $\alpha$ data points $F_\alpha$.  Notice
the difference to classical VB. In classical VB, we optimize the KL
divergence between $q(\cdot)$ and a posterior,
$\mathrm{KL}(q(\beta, \bz) || p(\beta, \bz \g \bx))$. Its objective is
a function of a fixed data set $\bx$; the objective in Eq.~\ref{eq:kl} is
a function of the population distribution $F_\alpha$.

We will use the mean-field variational family, where each latent
variable is independent and governed by a free parameter,
\begin{align}
  q(\beta, \bz) = q(\beta \g \lambda) \prod_{i=1}^{\alpha} q(z_i \g
  \phi_i).
  \label{eq:mean-field}
\end{align}
The free variational parameters are the global parameters $\lambda$
and local parameters $\phi_i$.  Though we focus on the mean-field
family, extensions could consider structured
families~\cite{saul_exploiting_1996, hoffman_structured_2015}, where
there is dependence between variables.

In classical VB, where we approximate the usual posterior, we cannot
compute the KL.  Thus, we optimize a proxy objective called the ELBO
(evidence lower bound) that is equal to the negative KL up to an
additive constant.  Maximizing the ELBO is equivalent to minimizing
the KL divergence to the posterior.

In population VB we also optimize a proxy objective, the F-ELBO.  The
F-ELBO is an expectation of the ELBO under the population distribution
of the data,
\begin{align}
  \cL(\lambda, \phi; F_\alpha) =
  \E_{F_\alpha} \left[
  \E_{q} \left[
  \log p(\beta) - \log q(\beta \g \lambda) +
  \sum_{i=1}^\alpha
  \log p(X_i, Z_i \g \beta) -
  \log q(Z_i)] \right] \right].
  \label{eq:felbo1}
\end{align}
The F-ELBO is a lower bound on the population evidence
$\log \E_{F_\alpha}[p(\bX)]$ and a lower bound on the negative KL to
the population posterior.  (See Appendix~A.)  The inner
expectation is over the latent variables $\beta$ and $\bZ$, and is a
function of the variational distribution $q(\cdot)$.  The outer
expectation is over the $\alpha$ random data points $\bX$, and is a
function of the population distribution $F_\alpha(\cdot)$.  The F-ELBO
is thus a function of both the variational distribution and the
population distribution.

As we mentioned, classical VB maximizes the (classical) ELBO, which is
equivalent to minimizing the KL.  The F-ELBO, in contrast, is only a
bound on the negative KL to the population posterior.  Thus maximizing 
the F-ELBO is suggestive but is not guaranteed to minimize the KL.
That said, our studies show that this is a good quantity to optimize
and in Appendix~A we show that the F-ELBO does
minimize $\E_{F_\alpha}[\mathrm{KL}(q(\bz, \beta) || p(\bz, \beta | \bX))]$. 
As we will see, the ability to sample from $F_\alpha$ is the only additional requirement for 
maximizing the F-ELBO, using stochastic gradients.

\parhead{Conditionally conjugate models.}  In the next section we will
develop a stochastic optimization algorithm to maximize
Eq.~\ref{eq:felbo1}.  First, we describe the class of models that we will
work with.

Following~\cite{hoffman_stochastic_2013} we focus on conditionally
conjugate models.  A conditionally conjugate model is one where each
complete conditional---the conditional distribution of a latent
variable given all the other latent variables and the
observations---is in the exponential family.  This class includes many
models in modern machine learning, such as mixture models, topic
models, many Bayesian nonparametric models, and some hierarchical
regression models.  Using conditionally conjugate models simplifies
many calculations in variational inference.

Under the joint in Eq.~\ref{eq:joint}, we can write a conditionally
conjugate model with two exponential families:
\begin{align}
  p(z_i, x_i\g\beta) & \textstyle = h(z_i, x_i) \exp\left \{
    \beta^\top t(z_i, x_i) - a(\beta)
  \right \} \label{eq:conjugate-0}\\
  p(\beta \g \zeta) & \textstyle = h(\beta) \exp\left \{ \zeta^\top
    t(\beta) - a(\zeta) \right \} \label{eq:conjugate}.
\end{align}
We overload notation for base measures $h(\cdot)$, sufficient
statistics $t(\cdot)$, and log normalizers $a(\cdot)$.  Note that
$\zeta$ is the hyperparameter and that
$t(\beta) = [\beta, -a(\beta)]$~\citep{bernardo_bayesian_2009}.

In conditionally conjugate models each complete conditional is in an
exponential family, and we use these families as the factors in the
variational distribution in Eq.~\ref{eq:mean-field}.  Thus $\lambda$
indexes the same family as $p(\beta \g \bz, \bx)$ and $\phi_i$ indexes
the same family as $p(z_i \g x_i, \beta)$.  For example, in latent
Dirichlet allocation~\citep{blei_latent_2003}, the complete
conditional of the topics is a Dirichlet; the complete conditional of
the per-document topic mixture is a Dirichlet; and the complete
conditional of the per-word topic assignment is a categorical.  (See
\cite{hoffman_stochastic_2013} for details.)

\parhead{Population variational Bayes.}  We have described the
ingredients of our problem.  We are given a conditionally conjugate
model, described in Eqs.~\ref{eq:conjugate-0}~and~\ref{eq:conjugate}, a
parameterized variational family in Eq.~\ref{eq:mean-field}, and a stream
of data from an unknown population distribution $F$.  Our goal is to
optimize the F-ELBO in Eq.~\ref{eq:felbo1} with respect to the
variational parameters.

The F-ELBO is a function of the population distribution, which is an
unknown quantity.  To overcome this hurdle, we will use the stream of
data from $F$ to form noisy gradients of the F-ELBO; we then update
the variational parameters with stochastic optimization.

Before describing the algorithm, however, we acknowledge one technical
detail.  Mirroring \cite{hoffman_stochastic_2013}, we optimize an
F-ELBO that is only a function of the global variational parameters.
The one-parameter population VI objective is
$\cL_{F_\alpha}(\lambda) = \max_{\phi} \cL_{F_\alpha}(\lambda, \phi)$.
This implicitly optimizes the local parameter as a function of the
global parameter.  The resulting objective is identical to
Eq.~\ref{eq:felbo1}, but with $\phi$ replaced by $\phi(\lambda)$.  (Details
are in Appendix~B).

The next step is to form a noisy gradient of the F-ELBO so that we can
use stochastic optimization to maximize it.  Stochastic optimization
maximizes an objective by following noisy and unbiased
gradients~\cite{robbins_stochastic_1951,bottou_online_1998}.  We will
write the gradient of the F-ELBO as an expectation with respect to
$F_\alpha$, and then use Monte Carlo estimates to form noisy
gradients.

We compute the gradient of the F-ELBO by bringing the gradient
operator inside the expectations of Eq.~\ref{eq:felbo1}.\footnote{For
  most models of interest, this is justified by the dominated
  convergence theorem.}  This results in a population expectation of
the classical VB gradient with $\alpha$ data points.

We take the natural gradient~\cite{amari_natural_1998}, which has a
simple form in completely conjugate
models~\citep{hoffman_stochastic_2013}.  Specifically, the natural
gradient of the F-ELBO is
\begin{align}
  \hat{\nabla}_\lambda \cL(\lambda; F_\alpha) =
  \E_{F_\alpha} \left[
  \zeta -
  \lambda +
  \sum_{i=1}^\alpha \E_{\phi_i(\lambda)}\left[t(x_i, Z_i)\right]
  \right].
  \label{eq:f-nat-grad}
\end{align}
We use this expression to compute noisy natural gradients at
$\lambda$. We collect $\alpha$ data points from $F$; for each we
compute the optimal local parameters $\phi_i(\lambda)$, which is a
function of the sampled data point and variational parameters; we then
compute the quantity inside the brackets in Eq.~\ref{eq:f-nat-grad}.  The
result is a single-sample Monte-Carlo estimate of the gradient, which
we can compute from a stream of data.  We follow the noisy gradient
and repeat.  This algorithm is summarized in Algorithm.~\ref{alg:felbo}. 
Since Equation~\ref{eq:f-nat-grad} is a Monte-Carlo estimate, we are free to draw $B$ data points from $F_\alpha$ 
(where $B << \alpha$) and rescale the sufficient statistics by $\alpha/B$. This makes the gradient estimate faster to calculate, but more noisy. 
As highlighted in Ref.~\cite{hoffman_stochastic_2013}, this makes sense because early iterations of the algorithm 
have inaccurate values of $\lambda$ so it is wasteful to pass through lots of data before making updates to $\lambda$.

\begin{algorithm}[t]
\caption{Population Variational Bayes}
\begin{algorithmic}
\label{alg:felbo}
\label{alg:pseudocode}
\STATE Randomly initialize global variational parameter $\lambda^{(0)}$
\STATE Set iteration $t \gets 0$ 
\REPEAT
\STATE Draw data minibatch $x_{1:B} \sim F_\alpha$
\STATE Optimize local variational parameters $\phi_1(\lambda^{(t)}),\dots,\phi_B(\lambda^{(t)})$
\STATE Calculate natural gradient $\hat{\nabla}_\lambda \cL(\lambda^{(t)}; F_\alpha)$ [see Eq.~\ref{eq:f-nat-grad}] 
\STATE Update global variational parameter with learning rate $\rho^{(t)}$\\ \setlength\parindent{24pt} $\lambda^{(t+1)} = \lambda^{(t)} + \rho^{(t)} \frac{\alpha}{B} \hat{\nabla}_\lambda \cL(\lambda^{(t)}; F_\alpha)$
\STATE Update iteration count $t \gets t + 1$
\UNTIL{forever}
\end{algorithmic}
\end{algorithm}

\parhead{Discussion.}  Thus far, we have defined the population
posterior and showed how to approximate it with population variational
inference.  Our derivation justifies using an algorithm like
stochastic variational inference (SVI)~\cite{hoffman_stochastic_2013}
on a stream of data.  It is nearly identical to SVI, but includes an
additional parameter: the number of data points in the population
posterior $\alpha$.

Note we can recover the original SVI algorithm as an instance of
population VI, thus reinterpreting it as minimizing the KL divergence
to the population posterior.  We recover SVI by setting $\alpha$ equal
to the number of data points in the data set and replacing the stream
of data $F$ with $\hat{F}_{\bx}$, the empirical distribution of the
observations.  The ``stream'' in this case comes from sampling with
replacement from $\hat{F}_{\bx}$, which results in precisely the
original SVI algorithm.\footnote{This derivation of SVI is an
  application of Efron's plug-in
  principle~\cite{efron_introduction_1994} applied to inference of the
  population posterior.  The plug-in principle says that we can
  replace the population $F$ with the empirical distribution of the
  data $\hat{F}$ to make population inferences.  In our empirical
  study, however, we found that population VB often outperforms
  SVI.  Treating the data in a true stream, and setting the
  number of data points different to the true number, can improve
  predictive accuracy.}

We focused on the conditionally conjugate family for convenience,
i.e., the simple gradient in Eq.~\ref{eq:f-nat-grad}.  We emphasize,
however, that by using recent tools for nonconjugate
inference~\citep{kingma_auto_2013, titsias_doubly_2014,
  ranganath_black_2014}, we can adapt the new ideas described
above---the population posterior and the F-ELBO---outside of
conditionally conjugate models.

\section{Empirical Evaluation}

We study the performance of population variational Bayes (population
VB) against SVI and SVB~\cite{broderick_streaming_2013}. With large
real-world data we study two models, latent Dirichlet
allocation~\cite{blei_latent_2003} and Bayesian nonparametric mixture
models, comparing the held-out predictive performance of the
algorithms. We study the data coming in a true ordered stream, and in
a permuted stream (to better match the assumptions of SVI). Across
data and models, population VB usually outperforms the existing
approaches.

\parhead{Models.} We study two models. The first is latent Dirichlet
allocation (LDA) \cite{blei_latent_2003}. LDA is a mixed-membership
model of text collections and is frequently used to find its latent
topics. LDA assumes that there are $K$ topics
$\beta_k \sim \textrm{Dir}(\eta)$, each of which is a multinomial
distribution over a fixed vocabulary. Documents are drawn by first
choosing a distribution over topics
$\theta_d \sim \textrm{Dir}(\alpha)$ and then drawing each word by
choosing a topic assignment $z_{dn} \sim \textrm{Mult}(\theta_d)$ and
finally choosing a word from the corresponding topic
$w_{dn} \sim \beta_{z_{dn}}$. The joint distribution is
\begin{align}
  p(\beta, \theta, \bz, \boldsymbol{w} | \eta, \gamma) = p(\beta | \eta) \prod_{d=1}^\alpha p(\theta_d | \gamma)
  \prod_{i=1}^N p(z_{di} | \theta_d) p(w_{di} | \beta, z_{di}).
\end{align}
Fixing hyperparameters, the inference problem is to estimate the
conditional distribution of the topics given a large collection of
documents.

The second model is a Dirichlet process (DP) mixture~\citep{escobar_bayesian_1995}.
Loosely, DP mixtures are mixture models with a potentially infinite
number of components; thus choosing the number of components is part
of the posterior inference problem. When using variational for DP
mixtures~\cite{blei_variational_2006}, we take advantage of the stick
breaking representation. The variables are mixture proportions
$\pi \sim \mathrm{Stick(\eta)}$, mixture components
$\beta_k \sim H(\gamma)$ (for infinite $k$), mixture assignments
$z_i \sim \mathrm{Mult}(\pi)$, and observations
$x_i \sim G(\beta_{z_i})$. The joint is
\begin{eqnarray}
  p(\beta, \pi, \bz, \bx | \eta, \gamma) =
  p(\pi | \eta) p(\beta | \gamma) \prod_{i=1}^\alpha p(z_{i} | \pi) p(x_{i} | \beta, z_{i}).
\end{eqnarray}
The likelihood and prior on the components are general to the
observations at hand. In our study of continuous data we use normal
priors and normal likelihoods; in our study of text data we use
Dirichlet priors and multinomial likelihoods.

For both models, we use $\alpha$, which corresponds to the number of data points in traditional analysis.

\begin{figure}[t]
  \centering
  \begin{subfigure}[b]{1.3\textwidth}
    \includegraphics[width=\textwidth]{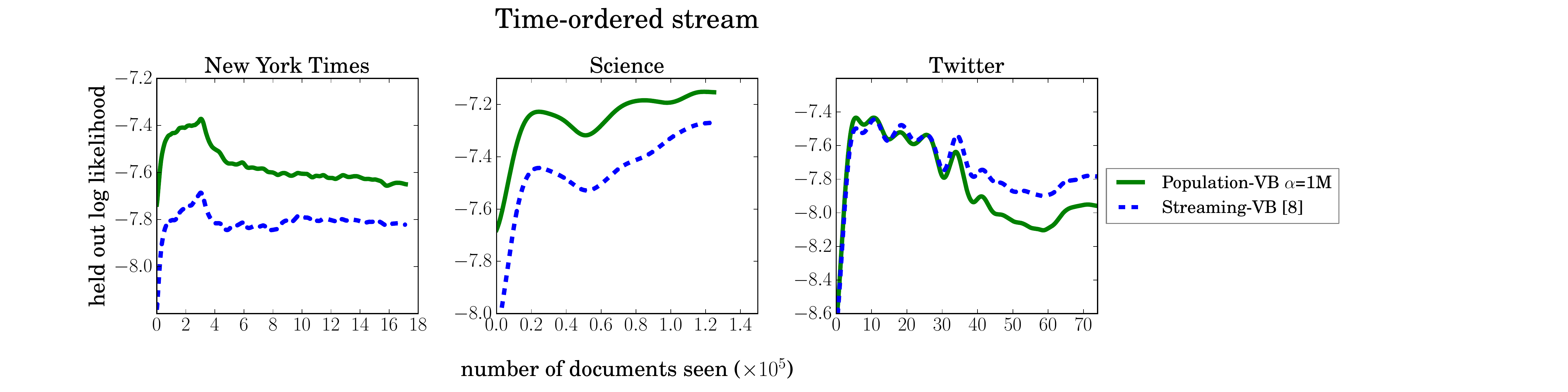}
    \label{fig:lda_time_order}
  \end{subfigure}
  \begin{subfigure}[b]{1.3\textwidth}
    \includegraphics[width=\textwidth]{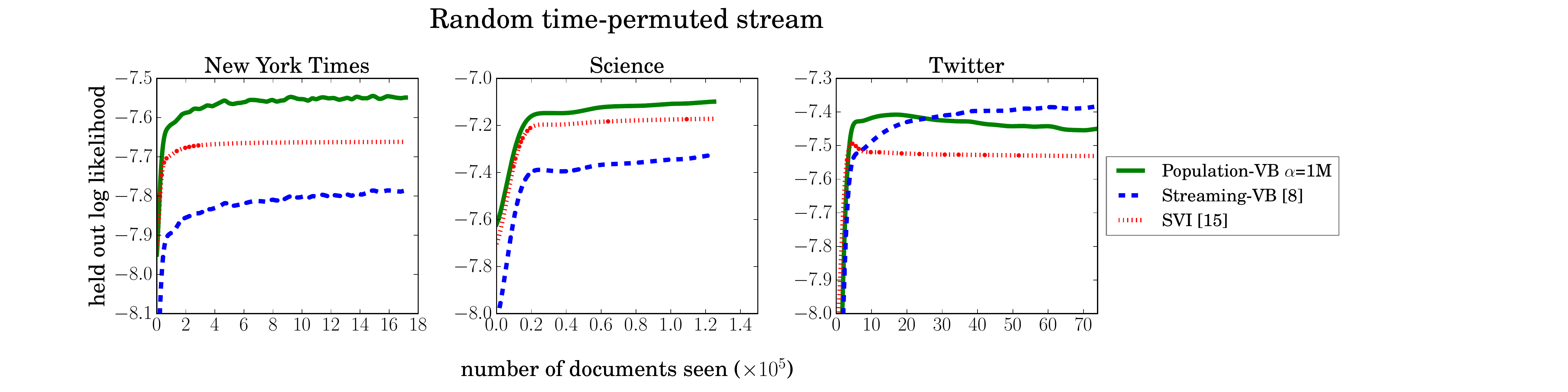}
    \label{fig:lda_time_shuffle}
  \end{subfigure}
  \vspace{-20pt}
  \caption{Held out predictive log likelihood for LDA on large-scale
    streamed text corpora. Population-VB outperforms existing
    methods for two out of the three settings. 
    We use the best settings of $\alpha$.}\label{fig:lda}
\end{figure}

\begin{figure}[h]
  \centering
  \begin{subfigure}[b]{1.3\textwidth}
    \includegraphics[width=\textwidth]{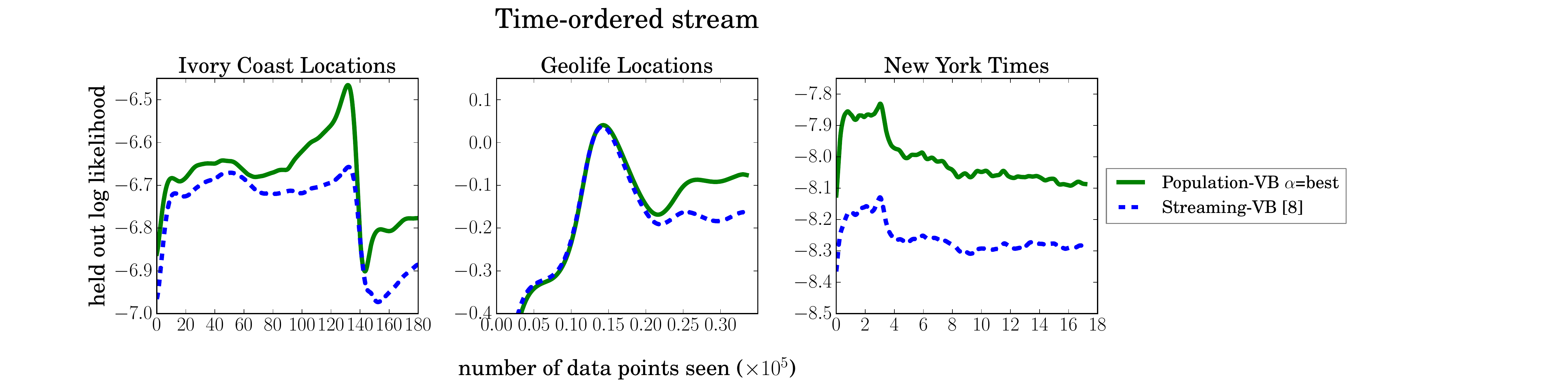}
    \label{fig:mix_time_order}
  \end{subfigure}
  \begin{subfigure}[b]{1.3\textwidth}
    \includegraphics[width=\textwidth]{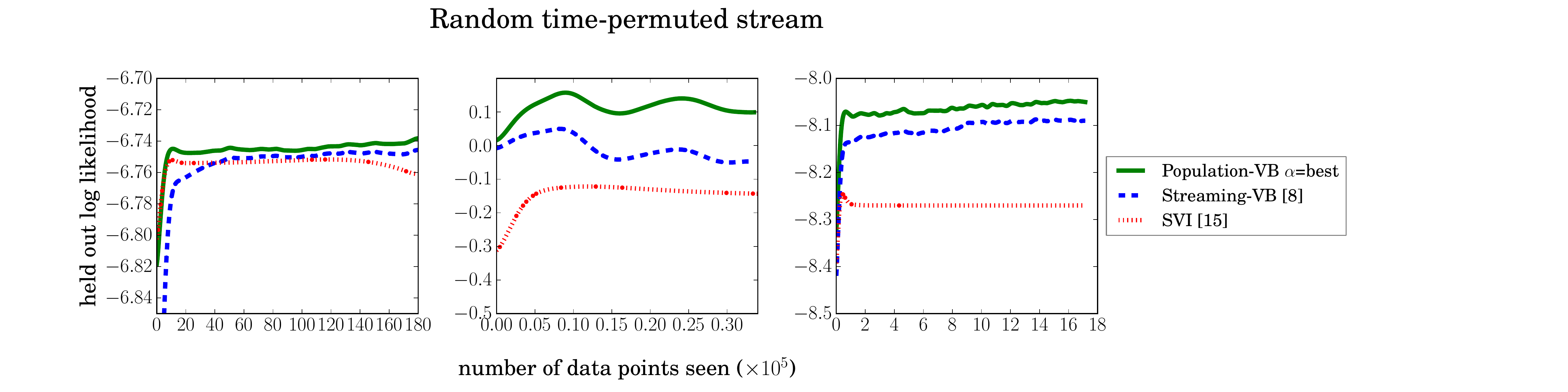}
    \label{fig:mix_time_shuffle}
  \end{subfigure}
  \vspace{-20pt}
  \caption{Held out predictive log likelihood for Dirichlet process
    mixture models on large-scale streamed location and text data
    sets. Note that we apply Gaussian likelihoods in the Geolife
    dataset, so the reported predictive performance is measured by
    probability density. We chose the best $\alpha$ for each population-VB curve}   
    \label{fig:mix}
\end{figure}

\begin{figure}[t]
  \centering

  \begin{subfigure}[c]{1.2\textwidth}
    \includegraphics[width=\textwidth]{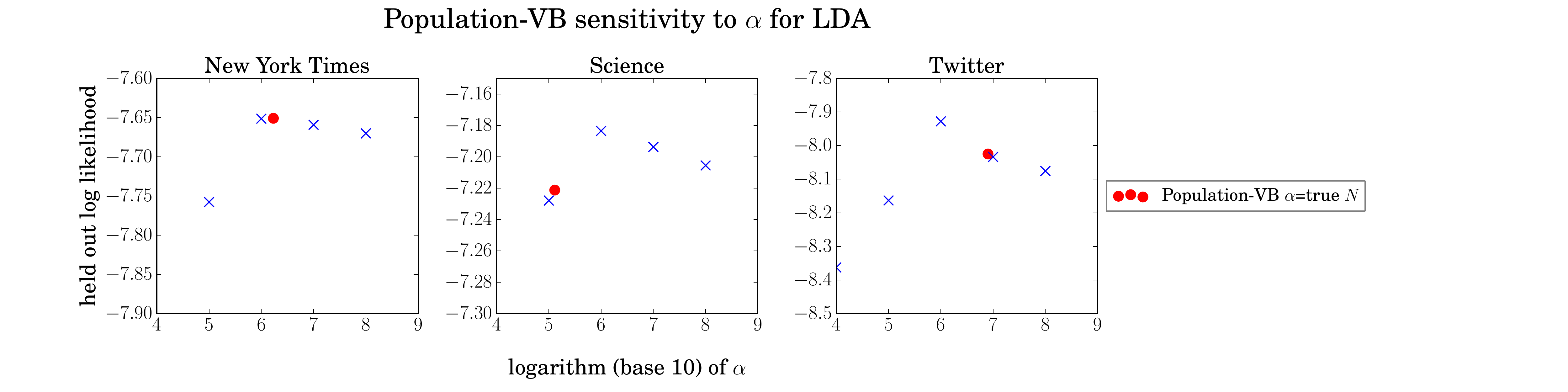}
    \label{fig:lda_sensitivity}
  \end{subfigure}
  \begin{subfigure}[c]{1.2\textwidth}
    \includegraphics[width=\textwidth]{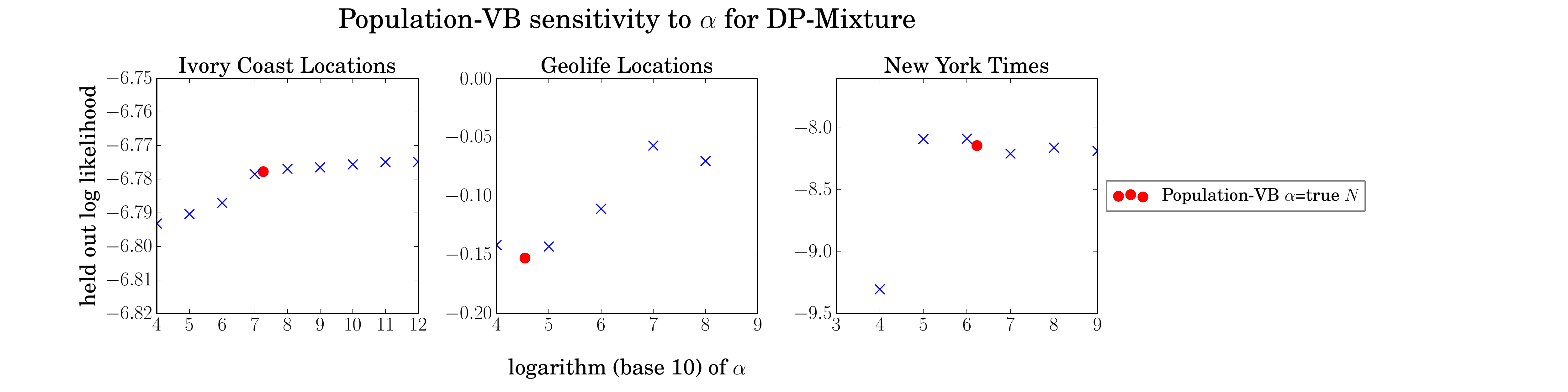}
    \label{fig:mix_sensitivity}
  \end{subfigure}
  \vspace{-10pt}
  \caption{We show the sensitivity of population-VB to
    hyperparameter $\alpha$ (based on final log likelihoods in
    the time-ordered stream) and find that the best setting of
    $\alpha$ often differs from the true number of data points
    (which may not be known in any case in practice).
  }\label{fig:sensitivity}
\end{figure}

\parhead{Datasets.} With LDA we analyze three large-scale streamed
corpora: 1.7M articles from the New York Times spanning 10 years, 130K
Science articles written over 100 years, and 7.4M tweets collected
from Twitter on Feb 2nd, 2014. We processed them all in a similar way,
choosing a vocabulary based on the most frequent words in the corpus
(with stop words removed): 8,000 for the New York Times, 5,855 for
Science, and 13,996 for Twitter. On Twitter, we each tweet is a
document, and we removed duplicate tweets and tweets that did not
contain at least 2 words in the vocabulary.

With DP mixtures, we analyze human location behavior data. These data
allow us to build periodic models\footnote{A simple account of periodicity is captured in the mixture model by observing the time of the week as one of the observation dimensions.} of human population mobility, with
applications to disaster response and urban planning. The Ivory Coast
location data contains 18M discrete cell tower locations for 500K
users recorded over 6~months~\citep{blondel_data_2012}. The Microsoft
Geolife dataset contains 35K latitude-longitude GPS locations for 182
users over 5 years. 
For both data sets, our
observations reflect down-sampling the data to ensure that each
individual is seen no more than once every 15 minutes.

\parhead{Results.} We compare population VB with SVI
\cite{hoffman_stochastic_2013} and streaming variational Bayes
(SVB)~\citep{broderick_streaming_2013} for
LDA~\citep{broderick_streaming_2013} and DP
mixtures~\citep{tank_streaming_2015}. SVB updates the variational
approximation of the global parameter using sequential Bayesian
updating, essentially accumulating expected sufficient statistics from
minibatches of data observed in a stream. (Here we give the final
results. We include the details of how we set and fit various
hyperparameters below.)

We measure model fitness by evaluating the average predictive log
likelihood on a set of held-out data. This involves splitting the
observations of held-out data points
into two equal halves, inferring the local component distribution
based on the first half, and testing with the second half \cite{hoffman_stochastic_2013}. 
For DP-mixtures, this works by predicting the location of a held-out data point by conditioning on the observed time of week.

In standard offline studies, the held-out set is randomly selected
from the data. With streams, however, we test on the next 10K
documents (for New York Times, Science), 500K tweets (for Twitter), or
25K locations (on Geo data). This is a valid held-out set because the
data ahead of the current position in the stream have not yet been
seen by the inference algorithms.

Figure~\ref{fig:lda} shows the performance of our algorithms for LDA.
We looked at two types of streams: one in which the data appear in
order and the other in which they have been permuted (i.e., an
exchangeable stream). The time permuted stream reveals performance
when each data minibatch is safely assumed to be an i.i.d. sample from
$F$; this results in smoother improvements to predictive likelihood.
On our data, we found that population VB outperformed SVI and SVB on
two of the data sets and outperformed SVI on all of the data. SVB
performed better than population VB on Twitter.

Figure~\ref{fig:mix} shows a similar study for DP mixtures. We
analyzed the human mobility data and the New York Times.
(Ref.~\cite{tank_streaming_2015} also analyzed the New York Times.) On
these data population VB outperformed SVB and SVI in all
settings.\footnote{Though our purpose is to compare algorithms, we
  make one note about a specific data set. The predictive accuracy for
  the Ivory Coast data set plummets after 14M data points.
  This is because of the data collection policy. For privacy
  reasons the data set provides the cell tower locations of a
  randomly selected cohort of 50K users every 2 weeks
  \citep{blondel_data_2012}. The new cohort at 14M data points behaves
  differently to previous cohorts in a way that affects predictive
  performance. However, both algorithms steadily improve after this
  shock.}

\parhead{Hyperparameters.} Our methods are based on stochastic
optimization and require setting the learning rate \citep{nemirovski_robust_2009}. For all
gradient-based procedures, we used a small fixed learning rate to
follow noisy gradients. We note that adaptive learning rates
\citep{duchi_adaptive_2011, tieleman_lecture_2012} are also applicable in this setting, though we did not observe an improvement using these for time-ordered streams.

Our procedures also require setting a batch size, how many data points
we observe before updating the approximate posterior. In the LDA study
we set the batch size to 100 documents for the larger corpora (New
York Times, Science) and 5,000 for Twitter. These sizes were selected
to make the average number of words per batch equal in both settings,
which helps lower the variance of the gradients. In the DP mixture
study we use a batch size of 5,000 locations for Ivory Coast, 500
locations for Geolife, and 100 documents for New York Times.

Unlike traditional Bayesian methods, the data set size $\alpha$ is a
hyperparameter to population VB. It helps control the posterior
variance of the population posterior. Figure~\ref{fig:sensitivity}
reports sensitivity to $\alpha$ for all studies (for the time-ordered
stream). These plots indicate that the optimal setting of $\alpha$ is
often different from the true number of data points; the best
performing population posterior variance is not necessarily the one
implied by the data.

LDA requires additional hyperparameters. In line with
\cite{hoffman_structured_2015}, we used 100 topics, and set the
hyperparameter to the global topics (which controls the word sparsity
of topics) to $\eta = 0.01$ and the hyperparameter to the word-topic
asssignments (which controls the sparsity of topic membership for each
word) to $\gamma = 0.1$. (We use these hyperparameters in
Eq.~\ref{eq:conjugate} and Eq.~\ref{eq:conjugate-0}.)
The DP mixture model requires a truncation hyperparameter $K$, which we set to 100 for all three data sets and verified that 
the number of components used after inference was less than this limit.

\section{Conclusions and Future Work}

We introduced a new approach to modeling through the population
posterior, a distribution over latent variables that combines
traditional Bayesian inference and with the frequentist idea of the
population distribution.  With this idea, we derived population
variational Bayes, an efficient algorithm for inference on streams.
On two complex Bayesian models and several large data sets, we found
that population variational Bayes usually performs better than
existing approaches to streaming inference.

%

\bibliographystyle{abbrv}
{\footnotesize
\bibliography{sigproc}}

\appendix

\section{Derivation and Bounds of the F-ELBO}
\label{app:derivation_felbo}
\label{app:kl}

Classic variational inference seeks to minimize $\mathrm{KL}( q(\beta, \bz) || p(\beta, \bz \g \bx))$ using the following equivalence to show that the negative evidence lower bound (ELBO) is an appropriate surrogate objective to be minimized,
\begin{eqnarray}
\log p(\bx) &=& \mathrm{KL}(q(\beta, \bz) || p(\bz \g \bx)) + \mathbb{E}_q [ \log p(\beta, \bz, \bx)  - \log q(\beta, \bz)] 
. \label{eq:basic_vi}
\end{eqnarray}
This equivalence arises from the definition of KL divergence \citep{wainwright_graphical_2008}. 

To derive the F-ELBO, replace $\bx$ with a draw $\bX$ of size $\alpha$ from the population distribution, $\bX \sim F_\alpha$, then apply an expectation with respect to $F_\alpha$ to both sides of Eq.\ref{eq:basic_vi},
\begin{eqnarray}
\E_{F_\alpha}[\log p(\bX)] &=& \E_{F_\alpha}[\mathrm{KL}(q(\beta, \bz) || p(\beta, \bz \g \bX)) + \mathbb{E}_q [ \log p(\beta, \bz, \bX)  - \log q(\beta, \bz)]] 
\nonumber \\
&=& \E_{F_\alpha}[\mathrm{KL}(q(\beta, \bz) || p(\beta, \bz \g \bX))] + \E_{F_\alpha}[\mathbb{E}_q [ \log p(\beta, \bz, \bX)  - \log q(\beta, \bz)]] 
. \label{eq:felbo_appendix}
\end{eqnarray}
This confirms that the negative F-ELBO is a surrogate objective for $\E_{F_\alpha}[\mathrm{KL}(q(\beta, \bz) || p(\beta, \bz \g \bX))]$ because $q(\cdot)$ does not appear on the left hand side of Eq.~\ref{eq:felbo_appendix}.

Now use the fact that logarithm is a concave function and apply Jensen's inequality to Eq.~\ref{eq:felbo_appendix} to show that the F-ELBO is a lower bound on the population evidence,
\begin{eqnarray}
\E_{F_\alpha}[\mathbb{E}_q [ \log p(\beta, \bz, \bX)  - \log q(\beta, \bz)]] &\le& \E_{F_\alpha}[\log p(\bX)]  \nonumber\\ &\le&\log \E_{F_\alpha}[p(\bX)]. \label{eq:felbo_ub}
\end{eqnarray}
Additionally, Jensen's inequality applied to Eq.~\ref{eq:felbo_appendix} in a different way shows that maximizing the F-ELBO minimizes an upper bound on the KL divergence between $q(\cdot)$ and the population posterior,
\begin{eqnarray}
\E_{F_\alpha}[\mathrm{KL}(q(\beta, \bz) || p(\beta, \bz \g \bX))] &=& \E_q[ \log q(\beta, \bz)] - \E_{F_\alpha}[ \E_q [\log p(\beta, \bz \g \bX)]] \nonumber \\
&\ge& \E_q[ \log q(\beta, \bz)] - \E_q[ \log \E_{F_\alpha}[p(\beta, \bz \g \bX)]] \nonumber \\
&=& \mathrm{KL}(q(\beta, \bz) || \E_{F_\alpha}[p(\beta, \bz \g \bX)]), \label{eq:felbo_lb}
\end{eqnarray}
where we have exchanged expectations with respect to $q(\cdot)$ and $F_\alpha$.

\section{One-Parameter F-ELBO}
\label{sec:one_parameter_felbo}

The F-ELBO for conditionally conjugate exponential families is as follows
\begin{align}
  \cL(\lambda, \phi; F_\alpha) =
  \E_{F_\alpha} \left[
  \E_{q} \left[
  \log p(\beta) - \log q(\beta \g \lambda) +
  \sum_{i=1}^\alpha
  \log p(X_i, Z_i \g \beta) -
  \log q(Z_i)] \right] \right].
  \nonumber
\end{align}
 This can be rewritten in terms of just the global variational
 parameters. We define the one parameter population variational
 inference objective as
 $\cL_{F_\alpha}(\lambda) = \max_{\phi} \cL_{F_\alpha}(\lambda,
 \phi)$.
 We can write this more compactly if we let $\phi_i(\lambda)$ be
 the value of $\phi_i$ that maximizes the F-ELBO given 
 $\lambda$.\footnote{The optimal local variational parameter $\phi_{i}$ can be computed using gradient ascent or coordinate 
 ascent as done in \cite{hoffman_stochastic_2013}.}
 Formally, this gives
 \begin{align*}
 \cL_{F_\alpha}(\lambda) =& \E_{q(\beta \g \lambda)} \left[ \log p(\beta) - \log q(\beta \g \lambda) + \E_{\bX \sim F_\alpha}  [\sum_{i=1}^\alpha \E_{q(Z_i \g \phi_{i}(\lambda))} \left[ \log p(X_i, Z_i \g \beta) - \log q(Z_i)] \right] \right],
 \end{align*}
where we have moved the expectation with respect to $F_\alpha$ inside the expectation with respect to $q(\cdot)$.

\end{document}